\documentclass[10pt,twocolumn,letterpaper]{article}

\usepackage{cvpr}              
\definecolor{cvprblue}{rgb}{0.21,0.49,0.74}
\usepackage[pagebackref,breaklinks,colorlinks,allcolors=cvprblue]{hyperref}
\usepackage[table]{xcolor} 
\usepackage{multirow}
\usepackage{arydshln}
\usepackage{float}
\usepackage{tcolorbox}

\def\paperID{454} 
\def\confName{CVPR}
\def\confYear{2026}

\title{Find Them All: Unveiling MLLMs for Versatile Person Re-identification}

\author{Jinhao Li\textsuperscript{\rm 1}, Zijian Chen\textsuperscript{\rm 2,4}, Lirong Deng\textsuperscript{\rm 3}, Guangtao Zhai\textsuperscript{\rm 2,4}\footnotemark[1], Changbo Wang\textsuperscript{\rm 1}\footnotemark[1] \\
\textsuperscript{\rm 1}School of Computer Science and Technology, East China Normal University \\
\textsuperscript{\rm 2}Institute of Image Communication and Information Processing, Shanghai Jiao Tong University, \\
\textsuperscript{\rm 3}Macao Polytechnic University, \textsuperscript{\rm 4}Shanghai AI Laboratory \\
{\tt\small lomljhoax@stu.ecnu.edu.cn, zijian.chen@sjtu.edu.cn, p2523406@mpu.edu.mo,} \\
{\tt\small zhaiguangtao@sjtu.edu.cn, cbwang@cs.ecnu.edu.cn} \\ 
}

\begin{document}%

\twocolumn[{%
\maketitle
\begin{figure}[H]
    \hsize=\textwidth
    \centering
    \includegraphics[width=\textwidth]{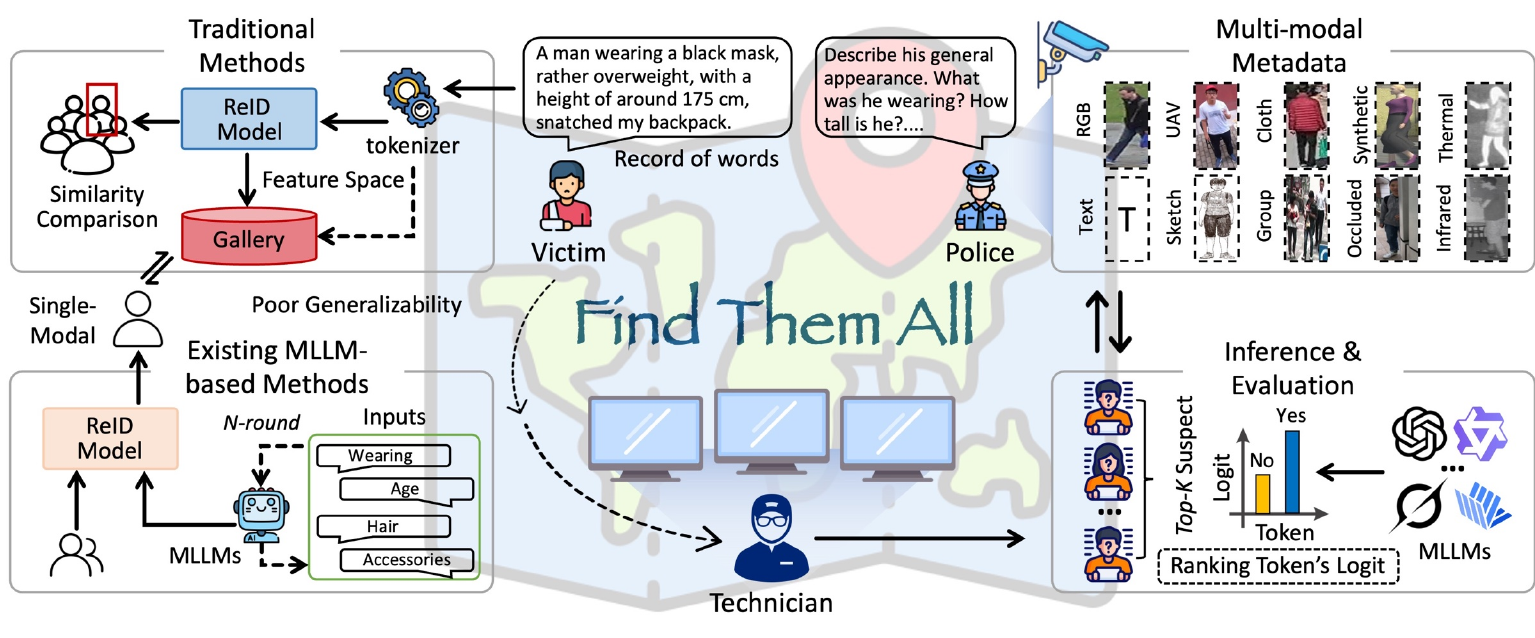}
    \caption{Comparison of different person re-identification (ReID) paradigms. Traditional methods typically matches a query, e.g., a suspect portrait or textual description, against a gallery by comparing high-level features. Similarly, existing MLLM-based methods utilize MLLMs to refine an initial description to gradually align with the target. However, both conventional paradigms predominantly concentrate on a few modalities (e.g., RGB and text), limiting their applicability in diverse real-world scenarios. In this work, we conduct a holistic evaluation for MLLMs on ten different modalities to verify their capabilities in handling person ReID tasks directly.}
    \label{fig:teaser}
\end{figure}
}]

\footnotetext[1]{Corresponding Authors.}

\maketitle
\begin{abstract}
Person re-identification (ReID) aims to retrieve images of a target person from the gallery set, with wide applications in medical rehabilitation and public security. However, traditional person ReID models are typically uni-modal, resulting in limited generalizability across heterogeneous data modalities. Recently, the emergence of multi-modal large language models (MLLMs) has shown a promising avenue for addressing this issue. Despite this potential, existing methods merely regard MLLMs as feature extractors or caption generators, leaving their capabilities in person ReID tasks largely unexplored. To bridge this gap, we introduce a novel benchmark for \underline{\textbf{V}}ersatile \underline{\textbf{P}}erson \underline{\textbf{Re}}-\underline{\textbf{ID}}entification, termed VP-ReID. The benchmark includes 257,310 multi-modal queries and gallery images, covering ten diverse person ReID tasks. In addition, we propose two task-oriented evaluation schemes for MLLM-based person ReID. Extensive experiments demonstrate the impressive versatility, effectiveness, and interpretability of MLLMs in various person ReID tasks. Nevertheless, they also have limitations in handling a few modalities, particularly thermal and infrared data. We hope that VP-ReID can facilitate the community in developing more robust and generalizable cross-modal foundation models for person ReID.
\end{abstract}    
\section{Introduction}

Person re-identification (ReID) aims to retrieve images of an interested person from the gallery images based on multi-modal queries \cite{ye2021deep}. The input modalities of person ReID vary significantly across different application scenarios, reflecting the diverse requirements of practical deployments. For instance, forensic applications often rely on sketch images drawn by artists and textual descriptions provided by witnesses, while intelligent surveillance systems may utilize thermal or infrared images to identify individuals under low-light conditions. These heterogeneous tasks necessitate the development of diverse input modalities, including RGB, sketch, synthetic, UAV, group, cloth-changing, occluded, thermal, and infrared images \cite{zhai2024multi, jiang2024domain, dai2025diffusion, zhang2024view, chen2024unsupervised, li2024disentangling, tan2024occluded, ling2023cross}, as well as textual descriptions \cite{jiang2025modeling}. Consequently, the primary challenge lies in designing a unified framework capable of effectively processing and integrating these diverse multi-modal inputs while maintaining robust performance across various real-world scenarios.

As shown in \cref{fig:teaser}, traditional person ReID methods \cite{huang2024federated, ye2023channel} predominantly focus on developing task-specific models that consist of a specialized tokenizer and retriever. The tokenizer is responsible for feature extraction, while the retriever handles the matching and ranking of candidate gallery images. Furthermore, these specialist components must be precisely paired to ensure optimal performance, meaning that a text tokenizer cannot be effectively paired with a sketch image retriever due to inherent modality disparities. While traditional person ReID approaches have achieved remarkable human-level performance within their respective modalities \cite{bao2023learning, zhang2022fmcnet}, they exhibit poor generalizability when transferred to other modalities. This fundamental lack of cross-modal generalizability significantly constrains the practical applicability of traditional person ReID methods in real-world scenarios, where diverse and heterogeneous input modalities are commonly encountered.

Recently, multi-modal large language models (MLLMs) have shown impressive performance across various multi-modal tasks \cite{chen2025just}. Pioneering works such as Flamingo \cite{alayrac2022flamingo} and LLaVA \cite{liu2023visual} exemplify the exceptional prowess of MLLMs as powerful visual-language learners. This characteristic holds significant potential for person ReID tasks, particularly in addressing the challenges of cross-modal inputs. Moreover, the inherent interpretability of MLLMs enables a clearer understanding of their decision-making processes. Existing MLLM-based person ReID methods can be categorized into two groups. One group utilizes MLLMs to recognize pedestrian image attributes or extract style features that capture human annotator preferences \cite{jiang2025modeling, wang2025idea}. The other group leverages MLLMs to generate rich textual descriptions for each pedestrian image directly \cite{zhai2024multi, he2024instruct, hu2024empowering}. However, these methods still rely on convolutional neural networks (CNN) or Transformer-based retrievers for person ReID, which fail to fully harness the comprehensive reasoning, instruction-following, and cross-modal understanding capabilities of MLLMs, thereby limiting their potential for truly unified multi-modal person ReID systems.

In this paper, we aim to enable MLLMs to directly and precisely retrieve the target individual from a gallery set, regardless of the query modality. To this end, we propose VP-ReID, a novel benchmark for versatile person ReID that encompasses two evaluation schemes: multiple-choice questions (MCQ) and query–gallery matching (QGM). The MCQ performs one-pass multi-choice reasoning, where all gallery images are jointly input for inference, while the QGM adopts a pairwise matching problem that involves multiple rounds of query–gallery inference. Both schemes include ten representative person ReID tasks, each accompanied by a corresponding dataset. In total, our VP-Bench comprises 4,642 query samples and 252,668 gallery images. Furthermore, we adopt a softmax-based evaluation pipeline for the QGM, which transforms the internal logits of output tokens into continuous and comparable matching scores. Finally, we conduct extensive evaluations of two traditional person ReID models, six proprietary MLLMs, and nine open-source MLLMs on our VP-ReID. The results provide an in-depth understanding of the strengths and weaknesses of MLLMs in person ReID tasks, establishing a comprehensive and unified foundation for future research. In conclusion, the contributions of this paper can be summarized as follows:
\begin{itemize}
    \item We construct VP-ReID, a novel benchmark for evaluating the person ReID capabilities of MLLMs, which contains 4,642 query samples and 252,668 gallery images across ten diverse person ReID tasks.
    \item We design two evaluation schemes for MLLM-based person ReID: multiple-choice questions (MCQ) and query–gallery matching (QGM), which systematically evaluate their strengths and weaknesses across various modality inputs and reasoning settings.
    \item We perform comprehensive evaluations of two traditional models, six proprietary MLLMs, and nine open-source MLLMs, offering an in-depth understanding of how MLLMs perform in person ReID tasks.
\end{itemize}
\begin{figure*}[t]
  \centering
  \includegraphics[width=\textwidth]{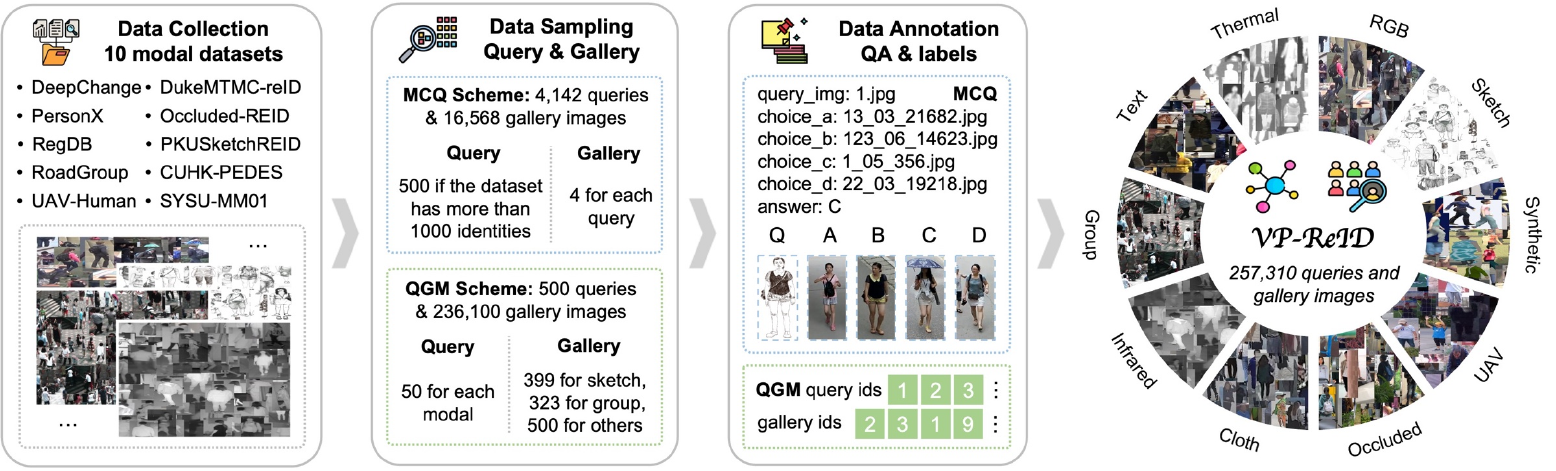}
  \caption{Construction pipeline of the {\bf VP-ReID}. We first collect source contents from ten person ReID datasets covering diverse modalities. Then, queries and gallery images are sampled based on the two proposed evaluation schemes: MCQ and QGM, resulting in a total of 257,310 queries and gallery images. Finally, we organize the data into question–answer pairs for the MCQ and extract labels for the QGM.}
  \label{fig:bench}
\end{figure*}


\section{Related Work}

\subsection{Person Re-identification}

Given a query image, the goal of person re-identification (ReID) is to retrieve the target image from a gallery set. Under the mainstream trend of deep learning, CNN-based person ReID methods are typically divided into two categories: closed-world and open-world \cite{ye2021deep}. The closed-world setting focuses on learning discriminative feature representations from well-labeled visible images captured by common video surveillance \cite{ye2024transformer}. In contrast, open-world ReID addresses more complex and challenging scenarios, such as cross-modal ReID \cite{yang2023towards, cheng2023unsupervised}, unsupervised learning \cite{dai2021idm, bai2021unsupervised}, and domain generalization \cite{li2021weperson, ni2022meta}. With the advent of the vision Transformer (ViT) \cite{dosovitskiy2020image}, a surge of ViT-based approaches has been proposed for person ReID, achieving remarkable performance on both regular image \cite{zhu2022dual, li2023dc} and cross-modal \cite{tan2024harnessing, rao2024hierarchical} ReID tasks. Moreover, ViT-based models have demonstrated clear superiority over CNN-based models in video person ReID \cite{wu2022cavit, wu2024temporal}. However, in the emerging era of MLLMs, the performance of MLLMs in person ReID remains unclear.

\subsection{MLLMs in Person Re-identification}

Recently, many works have utilized multi-modal large language models (MLLMs) for person ReID. For example, MP-ReID \cite{zhai2024multi} regards ChatGPT as a multi-prompt generator, which contributes to a comprehensive understanding of the input image. Similarly, Instruct-ReID \cite{he2024instruct} leverages MLLMs to generate instructions encompassing six traditional person ReID tasks. Though MLLMs reduce human labor and annotation time, the generated annotations lack diversity in description styles. To address this issue, TVI-LFM \cite{hu2024empowering} employs an off-the-shelf large language model to augment the generated textual descriptions from the vision language model (VLM), followed by an additional VLM that creates fusion features semantically consistent with visible features. Moreover, IDEA \cite{wang2025idea} incorporates MLLMs to extract eight predefined attributes from the generated captions and populates these attributes into the same template for more diverse descriptions. Most recently, HAM \cite{jiang2025modeling} performs clustering and prompt learning on the extracted textual descriptions, thereby enriching the diversity of the MLLM-generated captions. Nevertheless, these approaches primarily treat MLLMs as feature extractors or caption generators, failing to fully exploit their capabilities in person ReID tasks.
\section{The Proposed VP-ReID}


\subsection{Scheme Design}  

To comprehensively evaluate the multi-modal reasoning and retrieval capabilities of MLLMs in person ReID, we propose VP-ReID, a versatile person ReID benchmark that introduces two evaluation schemes: multi-choice questions (MCQ) and query-gallery matching (QGM). In the MCQ, all gallery images are fed into the model simultaneously for inference, enabling direct multi-choice reasoning within a single pass. The model is required to select the correct gallery image from the candidates given a query and a prompt, following a question-answer format. The prediction is considered correct only if the model selects the target gallery image, and the final performance is reported in terms of accuracy. In contrast, the QGM formulates person ReID as a binary matching problem between a query and a gallery image under prompt guidance, where the model outputs a single-word response (Yes or No) to indicate identity consistency. For each query, the model performs multiple rounds of inference, each comparing the query with one gallery image at a time. The evaluation of QGM follows the softmax-based evaluation pipeline described in \cref{eval}. Besides, both schemes contain ten different person ReID tasks, each accompanied by a corresponding dataset. This design allows the benchmark to systematically analyze the strengths and weaknesses of MLLMs in handling diverse modality inputs and reasoning settings.

\subsection{Construction Pipeline}

\noindent \textbf{Data Collection.} As shown in \cref{fig:bench}, the VP-ReID benchmark is established by identifying ten representative person ReID tasks, including RGB, sketch, synthetic, UAV, occluded, cloth-changing, group, text, thermal, and infrared person ReID. These tasks are selected for their strong relevance to real-world scenarios and their ability to expose the multi-modal limitations of existing models. Based on the defined tasks, we first investigate the research in these fields and adopt ten commonly used datasets for our experiments. The detailed statistics of these datasets and our VP-Bench are reported in \cref{tab:datasets}. 

\noindent \textbf{Data Sampling.} After data collection, we build the MCQ by sampling one query for each identity in these datasets. For datasets containing more than 1,000 identities, we select only 500 queries. The corresponding gallery set for each query is then composed of one positive image from the same identity and three negative images from the remaining identities. For datasets captured by multiple cameras, gallery samples are drawn from different cameras to enhance cross-view diversity. This process yields a total of 4,142 query images and 16,568 gallery images. 

For the QGM, we begin by selecting 50 distinct identities from each dataset to form the query set. The gallery set is subsequently assembled by pairing each query with one positive image from the same identity and 499 images from other identities (398 for sketch person ReID and 323 for group person ReID). Similar to the MCQ setting, gallery images are chosen from different cameras. Overall, the VP-ReID benchmark comprises 500 query samples and 236,100 gallery images for the QGM.


\begin{table}[!t]
    \centering
    \caption{Statistics of the used datasets and our VP-ReID. For CUHK-PEDES, RegDB, and SYSU-MM01, we report image/text, RGB/thermal, and RGB/infrared in the `\# Images' column, respectively. The `\# Query' and `\# Gallery' of the MCQ and QGM schemes are delimited by slashes.}
     \resizebox{1\linewidth}{!}{\begin{tabular}{lccccccc}
	\hline
	  Datasets  & \# Images & \# Identity & \# Query & \# Gallery & Type \\ 
    \hline
    DukeMTMC-ReID \cite{zheng2017unlabeled} & 36,441 & 1,812 & 702/50 & 4/500 &  RGB \\
    PKUSketchReID \cite{pang2018cross} & 400 & 200 & 200/50 & 4/399 & Sketch \\
    PersonX \cite{sun2019dissecting} & 45,576 & 1,266 & 856/50 & 4/500 & Synthetic \\
    UAV-Human \cite{li2021uav} & 41,290 & 1,144 & 119/50 & 4/500 & UAV \\
    Occluded-REID \cite{zhuo2018occluded} & 2,000 & 200 & 200/50 & 4/500 & Occluded \\
    DeepChange \cite{xu2023deepchange} & 178,000 & 1,121 & 500/50 & 4/500 & Cloth-chaning \\
    RoadGroup \cite{lin2019group} & 324 & 162 & 162/50 & 4/324 & Group \\
    CUHK-PEDES \cite{li2017person} & 40,206/80,422  & 13,003 & 500/50 & 4/500 & Text \\
    RegDB \cite{nguyen2017person} & 4,120/4,120 & 412 & 412/50 & 4/500 & Thermal \\
    SYSU-MM01 \cite{wu2017rgb}  & 20,284/9,929& 491 & 491/50 & 4/500 & Infrared \\
    \hline
    \end{tabular}}
    \vspace{-1em}
    \label{tab:datasets}
\end{table}

\noindent \textbf{Data Annotation.} Finally, we annotate the sampled data according to different evaluation schemes. In the MCQ, each instance is represented as a question–answer pair, primarily consisting of three components: (1) the path to the query image or query text, (2) the paths to the gallery images, and (3) an answer grounded in the data collection stage. The gallery images are shuffled to ensure randomness. For each gallery image, we assign a distinct capital letter so that we can prompt the MLLMs to respond solely with the corresponding letter, thereby facilitating straightforward accuracy computation. In the QGM, we extract the labels of the query samples and gallery images for the evaluation process. In both schemes, each query is paired with only one gallery image that shares the same identity.

\subsection{Evaluation Pipeline} 
\label{eval}

To establish a comparative bridge between traditional models and MLLMs in person ReID tasks, we adopt a softmax-based evaluation strategy \cite{wu2023q} to transform the output tokens of MLLMs into continuous and comparable matching scores. For each gallery image $g_j$ in the QGM, the MLLMs are required to provide a single-word response $x_j \in \{\textbf{\textit{Yes}},$ $\textbf{\textit{No}}\}$ reflecting their decision on identity consistency. By applying a softmax operation over the internal logits corresponding to the tokens \textbf{\textit{Yes}}($x_{j}^{Yes}$) and \textbf{\textit{No}}($x_{j}^{No}$), the matching confidence for the query–gallery pair is computed as:
\begin{equation}
p_{j} = \frac{e^{x_{j}^{Yes}}}{e^{x_{j}^{Yes}} + e^{x_{j}^{No}}}.
\end{equation}
For a given query $q_i$, with a gallery set size of $G$, we obtain a distance vector:
\begin{equation}
    s_i = [1-p_1^i, 1-p_2^i, ..., 1-p_G^i].
\end{equation}
By ranking the gallery images in ascending order of the scores in $s_i$, we can derive the traditional person ReID metrics, such as CMC and mAP:
\begin{equation}
    \text{CMC}(k) = \frac{1}{Q} \sum_{i=1}^{Q} \textbf{1}\big[\min_{g_j \in \mathcal{P}_i} \text{rank}(g_j) \leq k\big],
\end{equation}
\begin{equation}
    \text{mAP} = \frac{1}{Q} \sum_{i=1}^{Q} \text{AP}_i.
\end{equation}
where $Q$ is the number of queries, $\mathcal{P}_i$ is the set of correctly matched gallery images for query $q_i$, and $\text{AP}_i$ refers to the average precision for query $q_i$.

\section{Experiments}

\subsection{Experimental Setup}

\begin{table*}[t]
    \centering
    \caption{Performance ({\it Accuracy}) comparisons of evaluated MLLMs on ten person ReID tasks under the MCQ setting. We use the {\it instruct} version for all open-source MLLMs. The best and second-best results are in \textbf{bold} and \underline{underlined}, respectively.}
    \resizebox{1\linewidth}{!}{\begin{tabular}{l|ccccccccccc}
    \hline                 
        {\textbf{Model}}  & Cloth & RGB & Occluded & Synthetic & Sketch & Thermal & Text & Group & Infrared & UAV & {\it Avg.}  \\
        \hline
        \rowcolor{gray!20} \multicolumn{12}{l}{\textbf{Proprietary MLLMs:}} \\ 
        \hdashline
        Grok-2 \cite{Grok-2} & 0.27 & 0.26 & 0.33 & 0.29 & 0.46 & 0.25 & 0.25 & 0.21 & 0.25 & 0.29 & 0.29 \\
        Grok-4 \cite{Grok-4} & 0.29 & 0.25 & 0.17 & 0.30 & 0.66 & 0.29 & 0.23 & 0.21 & 0.35 & 0.24 & 0.30 \\
        Gemini-1.5-Pro \cite{team2024gemini} & 0.68 & 0.84 & \underline{0.98} & \underline{0.99} & 0.92 & 0.33 & \underline{0.97} & \textbf{0.98} & 0.51 & 0.76 & 0.80 \\
        Gemini-2.0-Flash \cite{gemini2} & 0.67 & 0.85 & 0.96 & 0.98 & \textbf{0.97} & \underline{0.46} & 0.95 & \textbf{0.98} & 0.53 & \underline{0.83} & 0.82 \\
        GPT-4o \cite{GPT-4o} & 0.71 & 0.86 & 0.97 & \underline{0.99} & 0.95 & 0.39 & \underline{0.97} & 0.94 & 0.49 & 0.63 & 0.79 \\
        GPT-4.1 \cite{GPT-4-1} & \textbf{0.79} & \textbf{0.92} &\textbf{0.99} & \textbf{1.00} & \underline{0.96} & \textbf{0.60} & \textbf{0.98} & \underline{0.96} & \textbf{0.63} & 0.76 & \textbf{0.86} \\
        \hline
        \rowcolor{gray!20} \multicolumn{12}{l}{\textbf{Open-Source MLLMs:}} \\ 
        \hdashline
        Qwen2.5-VL-3B \cite{bai2025qwen2} & 0.27 & 0.25 & 0.26 & 0.30 & 0.30 & 0.28 & 0.22 & 0.32 & 0.23 & 0.32 & 0.28 \\
        Qwen2.5-VL-7B \cite{bai2025qwen2} & 0.49 & 0.66 & 0.67 & 0.95 & 0.80 & 0.34 & 0.75 & 0.83 & 0.46 & 0.63 & 0.69 \\
        Qwen2.5-VL-32B \cite{bai2025qwen2} & 0.57 & 0.76 & 0.85 & 0.98 & 0.83 & 0.36 & 0.81 & 0.86 & 0.52 & 0.82 & 0.75 \\
        Qwen3-VL-2B \cite{yang2025qwen3} & 0.56 & 0.62 & 0.76 & 0.85 & 0.77 & 0.35 & 0.68 & 0.72 & 0.45 & 0.63 & 0.66 \\
        Qwen3-VL-4B \cite{yang2025qwen3} & 0.71 & 0.83 & 0.89 & \underline{0.99} & 0.90 & 0.39 & 0.90 & \underline{0.96} & 0.53 & 0.71 & 0.82 \\
        Qwen3-VL-8B \cite{yang2025qwen3} & 0.71 & 0.88 & 0.97 & \underline{0.99} & 0.88 & \underline{0.46} & 0.90 & \underline{0.96} & \underline{0.59} & \textbf{0.85} & \underline{0.84} \\
        Qwen3-VL-32B \cite{yang2025qwen3} & 0.68 & \underline{0.89} & 0.96 & 0.98 & 0.84 & 0.40 & 0.91 & \textbf{0.98} & 0.56 & 0.82 & 0.83 \\
        InternVL3.5-8B \cite{wang2025internvl3_5} & 0.61 & 0.65 & 0.96 & 0.95 & 0.74 & 0.30 & 0.26 & 0.84 & 0.39 & 0.65 & 0.66 \\
        InternVL3.5-38B \cite{wang2025internvl3_5} & \underline{0.75} & 0.75 & 0.85 & \textbf{1.00} & 0.88 & 0.36 & 0.25 & \textbf{0.98} & 0.58 & 0.81 & 0.72 \\
        \hline
    \end{tabular}}
    \label{tab:mcq}
\end{table*}

\begin{table*}[t]
    \centering
    \caption{Performance ({\it mAP}) comparisons between traditional models and MLLMs on ten person ReID tasks under the QGM setting. For the `Text' column, we use IRRA \cite{jiang2023cross} as the traditional baseline and obtain an mAP of 0.59. We use the {\it instruct} version for all evaluated MLLMs. The best and second-best results are in \textbf{bold} and \underline{underlined}, respectively.}
    \resizebox{1\linewidth}{!}{\begin{tabular}{l|ccccccccccc}
    \hline                 
        {\textbf{Model}}  & Cloth & RGB & Occluded & Synthetic & Sketch & Thermal & Text & Group & Infrared & UAV & {\it Avg.}  \\
        \hline
        \rowcolor{gray!20} \multicolumn{12}{l}{\textbf{Traditional Models:}} \\ 
        \hdashline
        TransReID \cite{he2021transreid} & \textbf{0.52} & \textbf{0.73} & \textbf{0.77} & \underline{0.95} & 0.16 & 0.02 & \textendash & 0.56 & 0.03 & \textbf{0.60} & 0.48 \\
        \hline
        \rowcolor{gray!20} \multicolumn{12}{l}{\textbf{Open-Source MLLMs:}} \\ \hdashline
        Qwen2.5-VL-3B \cite{bai2025qwen2} & 0.03 & 0.06 & 0.11 & 0.43 & 0.22 & 0.02 & 0.45 & 0.12 & 0.07 & 0.21 & 0.20 \\
        Qwen2.5-VL-7B \cite{bai2025qwen2} & 0.25 & 0.47 & 0.37 & 0.60 & 0.39 & \underline{0.08} & 0.55 & 0.56 & 0.11 & 0.33 & 0.44 \\
        Qwen2.5-VL-32B \cite{bai2025qwen2} & 0.22 & 0.45 & 0.23 & 0.77 & 0.37 & 0.03 & 0.55 & 0.57 & 0.06 & 0.31 & 0.43 \\
        Qwen3-VL-2B \cite{yang2025qwen3} & 0.33 & 0.43 & 0.46 & 0.87 & 0.48 & \textbf{0.09} & 0.58 & 0.65 & 0.07 & 0.37 & 0.52 \\
        Qwen3-VL-4B \cite{yang2025qwen3} & \underline{0.49} & \underline{0.66} & \underline{0.74} & \textbf{0.99} & 0.60 & 0.04 & \underline{0.74} & 0.80 & 0.14 & 0.43 & \underline{0.68} \\
        Qwen3-VL-8B \cite{yang2025qwen3} & \underline{0.49} & 0.57 & 0.63 & \underline{0.95} & \underline{0.77} & \textbf{0.09} & 0.68 & \underline{0.84} & \textbf{0.17} & \underline{0.54} & \underline{0.68} \\
        Qwen3-VL-32B \cite{yang2025qwen3} & 0.46 & 0.57 & 0.63 & 0.89 & \textbf{0.78} & \textbf{0.09} & \textbf{0.83} & \textbf{0.92} & \underline{0.16} & 0.43 & \textbf{0.69} \\
        InternVL3.5-8B \cite{wang2025internvl3_5} & 0.22 & 0.04 & 0.66 & 0.28 & 0.10 & 0.02 & 0.01 & 0.05 & 0.02 & 0.04 & 0.18 \\
        InternVL3.5-38B \cite{wang2025internvl3_5} & 0.31 & 0.47 & 0.66 & 0.98 & 0.32 & 0.03 & 0.00 & 0.15 & 0.06 & 0.22 & 0.39 \\
        \hline
    \end{tabular}}
    \label{tab:performance}
\end{table*}

\begin{figure*}[t]
  \centering
  \includegraphics[width=\textwidth]{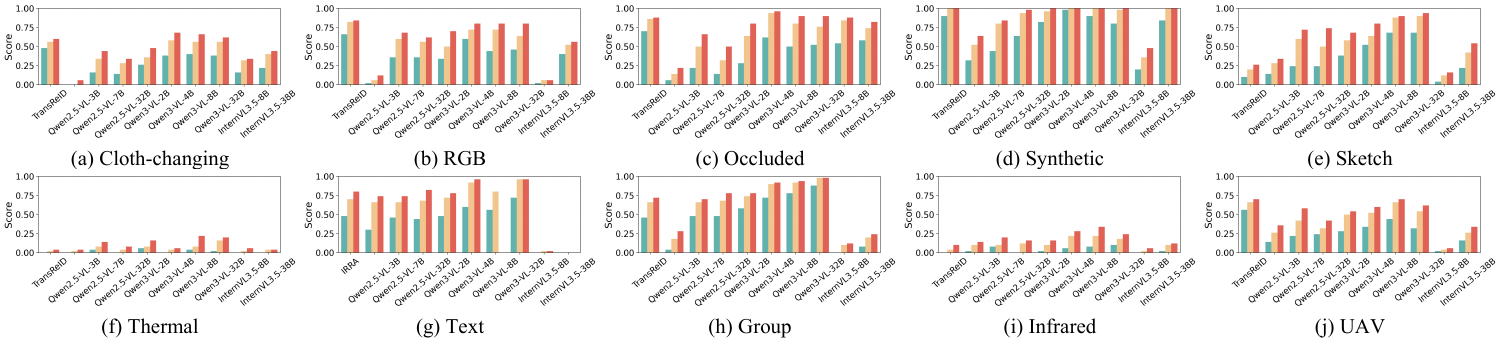}
  \caption{Performance ({\it CMC}) comparisons between traditional models and MLLMs on ten person ReID tasks under the QGM setting.}
  \label{fig:rank}
\end{figure*}

Our experiments evaluate a total of 15 MLLMs, encompassing a diverse mix of proprietary and open-source models. For proprietary models, we adopt OpenAI models (GPT-4o \cite{GPT-4o} and GPT-4.1 \cite{GPT-4-1}), Google's Gemini models (Gemini 2.0 Flash \cite{gemini2} and Gemini 1.5 Pro \cite{team2024gemini}), and Grok models (Grok-2 \cite{Grok-2} and Grok-4 \cite{Grok-4}). For open-source models, we cover the Qwen2.5-VL \cite{bai2025qwen2}, Qwen3-VL \cite{yang2025qwen3}, and InternVL families, including InternVL2.5 \cite{chen2024expanding}, InternVL3 \cite{zhu2025internvl3}, and InternVL3.5 \cite{wang2025internvl3_5}. To enable a more comprehensive comparison, we also include representative traditional models. Specifically, we adopt TransReID \cite{he2021transreid}, a transformer-based object ReID framework that achieves strong performance across various benchmarks. For the text person ReID, we employ IRRA \cite{jiang2023cross}, a cross-modal ReID framework that models the relationships between local visual and textual tokens while enhancing global image-text matching without relying on additional prior supervision. All traditional models are reproduced using their publicly available code.  



\subsection{Main Results}

\noindent \textbf{Results on the MCQ.} As shown in \cref{tab:mcq}, proprietary MLLMs outperform open-source counterparts, with GPT-4.1 achieving the highest average accuracy of 0.86. Notably, GPT-4.1 attains 0.60 and 0.63 on the thermal and infrared person ReID tasks, exceeding the second-best results by 0.14. Meanwhile, open-source MLLMs also demonstrate competitive performance. For example, Qwen3-VL-8B and Qwen3-VL-32B achieve the best results on UAV and group person ReID tasks, respectively. These results offer an initial overview of the MLLMs' performance on person ReID tasks. To enable a comprehensive comparison between traditional models and MLLMs, we focus on the QGM evaluation scheme for further analysis and discussion in the following experiments.

\noindent \textbf{Results on the QGM.} The performance of MLLMs on the QGM is reported in \cref{tab:performance} and \cref{fig:rank}. On the one hand, several MLLMs (e.g., Qwen3-VL family) obtain remarkable results on tasks such as RGB, occluded, synthetic, sketch, text, and group person ReID. Notably, Qwen3-VL-32B achieves an impressive mAP of 0.83 on the text person ReID task and 0.92 on the group person ReID task, effectively retrieving nearly all target images without any training data. On the other hand, MLLMs also show substantial limitations in VP-ReID. For example, they struggle significantly with thermal and infrared person ReID tasks. Quantitatively, even the best-performing model attains only 0.09 and 0.17 in mAP on thermal and infrared tasks, respectively. This degradation in performance may stem from the inherent information loss in these modalities, as well as from the absence of domain-specific training data for MLLMs. 

\noindent \textbf{Comparisons with Traditional Person ReID Models.} To ensure a more comprehensive comparison, we adopt two traditional person ReID models, TransReID and IRRA, which were trained on Market-1501 \cite{zheng2015scalable} and ICFG-PEDES \cite{ding2021semantically}, respectively. In contrast, the MLLMs are evaluated in a zero-shot setting. We present the results in \cref{tab:performance} and \cref{fig:rank}. It can be observed that TransReID achieves state-of-the-art results on cloth-changing, RGB, occluded, and UAV person ReID tasks. However, the performance gap between the MLLMs and TransReID is relatively small. In particular, the difference in mAP is less than 0.03 in the cloth-changing and occluded person ReID tasks. Moreover, the MLLMs outperform TransReID by a large margin in sketch and group person ReID tasks, highlighting their superior cross-modal capabilities. Similarly, Qwen3-VL-32B surpasses IRRA by 0.24 in Rank@1 and mAP on the text person ReID task. These results indicate that, while traditional models perform well on specific modalities, MLLMs can achieve comparable performance in these modalities and significantly outperform them in others.

\begin{table}[t]
    \setlength{\tabcolsep}{2pt}
    \centering
    \caption{Results on grayscale RGB and group person ReID tasks. Rank@1, Rank@5, Rank@10, and mAP are delimited by slashes. The best and second-best results of Rank@1 and mAP are in \textbf{bold} and \underline{underlined}, respectively.}
    \fontsize{9}{11}\selectfont 
    \resizebox{1\linewidth}{!}{\begin{tabular}{l|cc}
    \hline                 
    {\textbf{Model}} & RGB & Group \\
    \hline
    \multicolumn{3}{l}{\textbf{Traditional Models:}} \\ 
    \hline
    \rowcolor{gray!20} TransReID & 0.24/0.42/0.50/0.33 & 0.16/0.28/0.34/0.23 \\
    \hline
    \multicolumn{3}{l}{\textbf{Open-Source MLLMs:}} \\ 
    \hdashline
    Qwen2.5-VL-3B & 0.02/0.10/0.12/0.05 & 0.04/0.14/0.18/0.09 \\
    Qwen2.5-VL-7B & 0.16/0.32/0.38/0.23 & 0.20/0.36/0.52/0.29 \\
    Qwen2.5-VL-32B & 0.12/0.30/0.34/0.20 & 0.28/0.54/0.60/0.39 \\
    Qwen3-VL-2B & \underline{0.28}/0.40/0.48/0.35 & 0.28/0.38/0.50/0.36 \\
    Qwen3-VL-4B & 0.22/0.40/0.52/0.33 & 0.46/0.70/0.78/0.57 \\
    Qwen3-VL-8B & 0.26/0.46/0.58/\underline{0.37} & \underline{0.56}/0.80/0.82/\underline{0.64} \\
    Qwen3-VL-32B & \textbf{0.38}/0.46/0.52/\textbf{0.42} & \textbf{0.66}/0.86/0.90/\textbf{0.76} \\
    InternVL3.5-8B & 0.00/0.02/0.02/0.01 & 0.02/0.02/0.06/0.04 \\
    InternVL3.5-38B & 0.08/0.18/0.22/0.13 & 0.04/0.10/0.12/0.08 \\
    \hline
    \end{tabular}}
    \label{tab:grayscale}
\end{table}

\begin{table*}[t]
    \setlength{\tabcolsep}{5pt}
    \caption{Comparisons of IRRA, DiaNA, Qwen3-VL-2B, and Qwen3-VL-4B on chat-based person ReID task. The best and second-best results of Rank@1 and mAP are in \textbf{bold} and \underline{underlined}, respectively}
    \label{tab:chat}
    \centering 
    \begin{tabular}{ccccc}
	\hline
	Models & IRRA \cite{jiang2023cross} & DiaNA \cite{bai2025chat} & Qwen3-VL-2B \cite{yang2025qwen3} & Qwen3-VL-4B \cite{yang2025qwen3} \\
    \hline
    Rank@1/@5/@10/mAP & 0.48/0.68/0.76/0.45 & \textbf{0.76}/\textbf{0.91}/\textbf{0.95}/\textbf{0.67} & 0.44/0.62/0.68/0.41 & \underline{0.52}/\underline{0.69}/\underline{0.77}/\underline{0.49} \\
    \hline
    \end{tabular}
\end{table*}

\noindent \textbf{MLLMs Perform Poorly on Thermal and Infrared Modalities.} To explore the underlying causes of MLLMs' poor performance on thermal and infrared person ReID tasks, we conduct additional experiments on RGB and group ReID tasks, where the MLLMs perform relatively well. Specifically, all images are converted to grayscale to eliminate the influence of color information. As reported in \cref{tab:grayscale}, the results show a moderate decline in performance compared with the original setting. However, most MLLMs still maintain high performance. Notably, Qwen3-VL-32B achieves 0.66 in Rank@1 and 0.76 in mAP on the group ReID task. These findings suggest that the absence of color information is not the primary factor contributing to the poor performance of MLLMs on thermal and infrared person ReID tasks. Instead, the main limitation lies in the visual encoders of MLLMs, which have not been sufficiently trained on large-scale thermal and infrared datasets.

\noindent \textbf{Chat-based Person ReID.} Recently, the person ReID research community has witnessed a surge of interest in chat-based person ReID methods. Therefore, we also compare the performance of these approaches with that of MLLMs. The evaluation is conducted on the test set of the ChatPedes \cite{bai2025chat} dataset, where the chats are first transformed into descriptive sentences by Qwen3-VL-8B and are subsequently fed into the MLLMs for inference. It is worth noting that IRRA and DiaNA are trained on the training set of the ChatPedes dataset, while the MLLMs operate in a completely training-free manner. Despite this, the results in \cref{tab:chat} show that Qwen3-VL-2B obtains 0.44 on Rank@1 and 0.41 on mAP, reaching competitive performance with IRRA. Moreover, Qwen3-VL-4B achieves further gains, indicating a strong potential to approach state-of-the-art performance when trained on ChatPedes.

\subsection{In-Depth Analyses}

\begin{figure}[t]
  \centering
  \includegraphics[width=\linewidth]{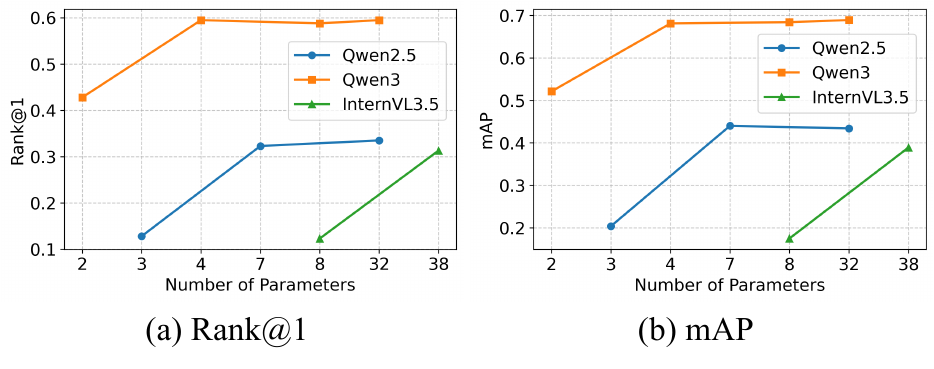}
  \caption{Scaling behavior of different model families on our VP-ReID. We report their averaged Rank@1 and mAP on eight tasks.}
  \label{fig:scale}
\end{figure}

\begin{table*}[t]
    \setlength{\tabcolsep}{4pt}
    \caption{Comparisons of Rank@1, mAP, and per-query inference time of Qwen3-VL-2B and Qwen3-VL-4B on RGB, text, group person ReID tasks. The performance before and after deployment with vLLM are delimited by slashes.}
    \label{tab:speed}
    \centering 
    \resizebox{1\linewidth}{!}{\begin{tabular}{cccccccccc}
    \hline
    \multirow{2}{*}{Models} & \multicolumn{3}{c}{RGB} & \multicolumn{3}{c}{Text} & \multicolumn{3}{c}{Group} \\
    \cline{2-4} \cline{5-7} \cline{8-10}
     & Rank@1 & mAP & Time(s) & Rank@1 & mAP & Time(s) & Rank@1 & mAP & Time(s) \\
    \hline
    Qwen3-VL-2B \cite{yang2025qwen3} & 0.34/0.34 & 0.43/0.44 & 37.6/4.88 & 0.48/0.44 & 0.58/0.56 & 59.1/6.44 & 0.58/0.58 & 0.65/0.65 & 27.3/3.48 \\
    Qwen3-VL-4B \cite{yang2025qwen3} & 0.60/0.58 & 0.66/0.66 & 43.2/5.64 & 0.60/0.66 & 0.74/0.78 & 62.5/8.00 & 0.72/0.72 & 0.80/0.80 & 31.2/4.24 \\
    \hline
    \end{tabular}}
\end{table*}

\noindent \textbf{Scaling Behavior.} To exemplify how MLLMs benefit from scaling laws on person ReID tasks, we averaged their Rank@1 and mAP on our VP-ReID (excluding thermal and infrared person ReID) and visualized the results in \cref{fig:scale}. The overall averaged performance adheres to the expected scaling trend, underscoring the potential of scaling MLLMs for improving person ReID capabilities. Nevertheless, we also notice that larger models within the same family do not always outperform their smaller counterparts, which may be caused by the substantial modality gap between these and RGB-based tasks. As shown in \cref{tab:performance}, Qwen3-VL-8B performs worse than Qwen3-VL-4B on cloth-changing and text person ReID tasks. A similar trend can be observed in the Qwen2.5-VL family, where Qwen2.5-VL-32B does not consistently surpass Qwen2.5-VL-7B.

\begin{figure}[t]
  \centering
  \includegraphics[width=\linewidth]{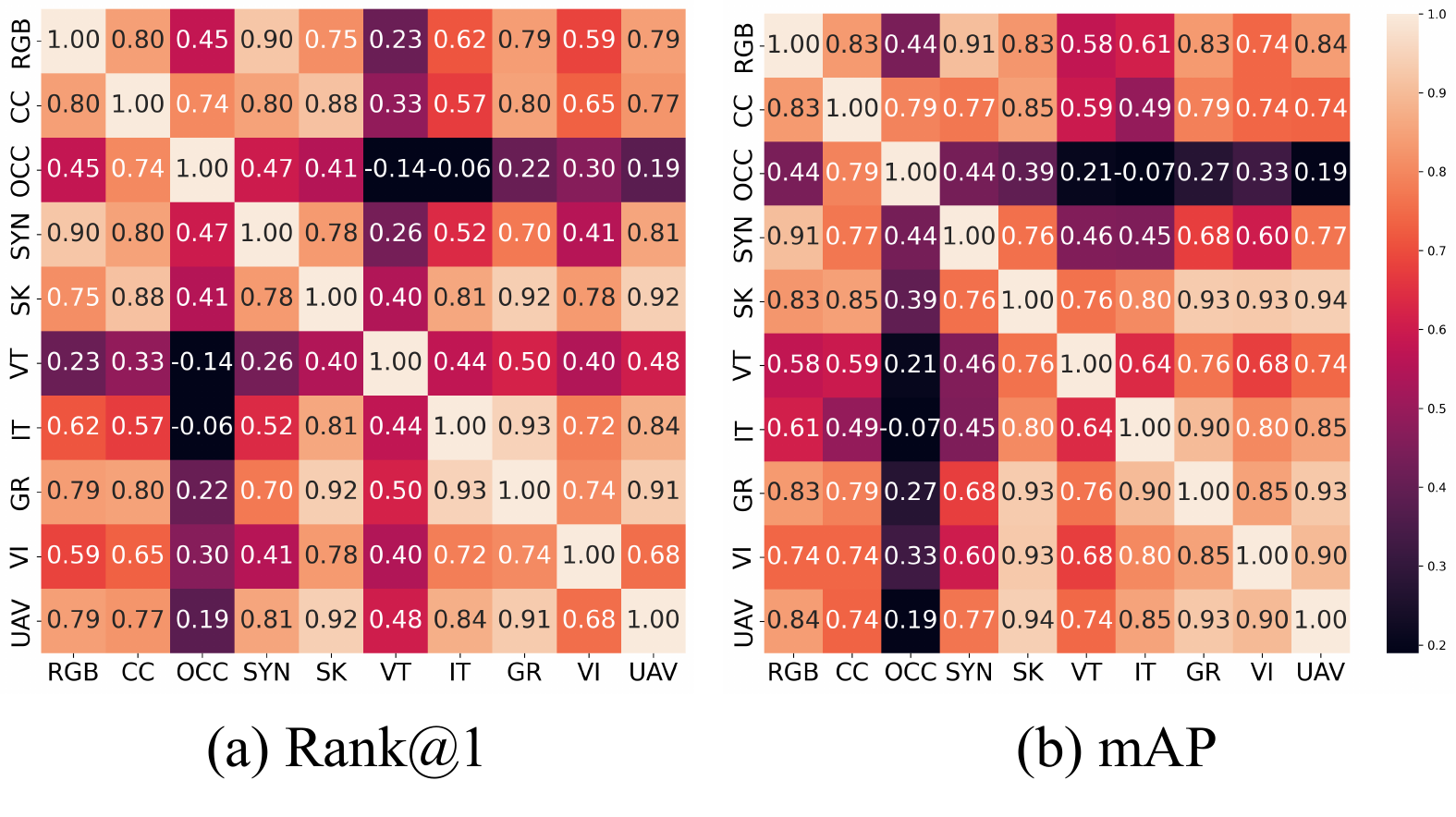}
  \caption{Correlation across different tasks in VP-ReID.} 
  \label{fig:correlation}
\end{figure}

\noindent \textbf{Correlation Analysis.} We present an analysis of the correlation among different person ReID tasks. Specifically, we compute the Pearson correlation coefficient among all tasks in VP-ReID. The results are illustrated in \cref{fig:correlation}. It can be observed that the RGB, cloth-changing, synthetic, sketch, group, and UAV person ReID tasks all correlate with one another. Interestingly, the MLLMs exhibit a strong performance correlation between sketch and other person ReID tasks, which may be attributed to substantial exposure to sketch-related data during their training process. In contrast, occluded, thermal, text, and infrared person ReID tasks correlate relatively weakly with all other tasks. That is because these tasks predominantly involve cross-modal scenarios or exhibit a substantial modality gap compared to the standard RGB person ReID task. These findings suggest that the future development of MLLM-based person ReID methods should prioritize addressing these disparate modalities rather than focusing on strongly correlated ones, thereby advancing the field toward a more comprehensive cross-modal understanding.

\begin{figure}[t]
  \centering
  \includegraphics[width=\linewidth]{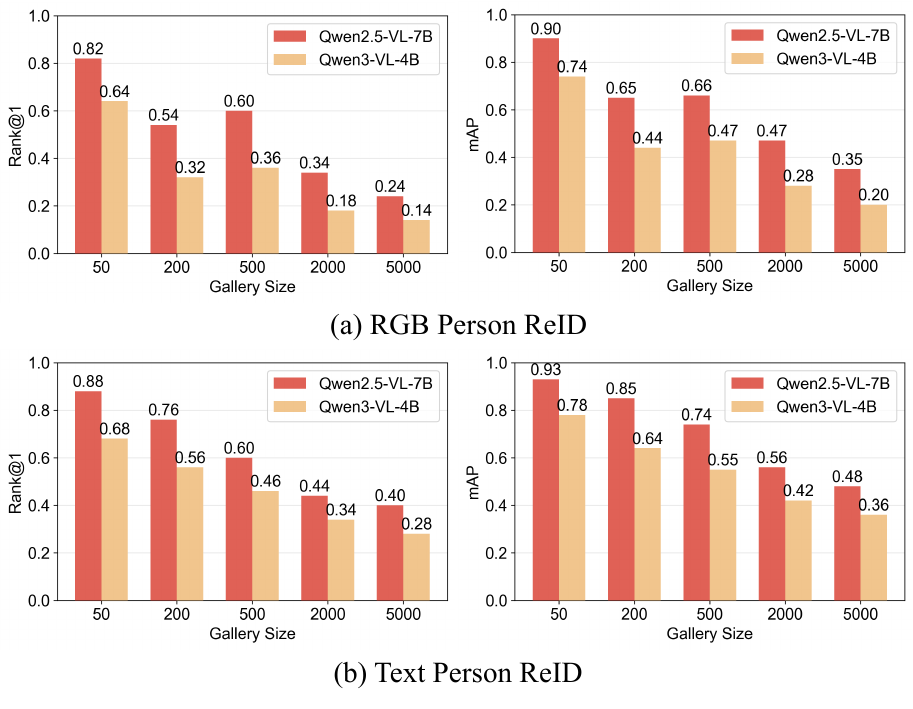}
  \caption{Impact of gallery size on the Rank@1 and mAP of Qwen2.5-VL-7B and Qwen3-VL-4B.}
  \label{fig:bar}
\end{figure}

\noindent \textbf{Inference Speed and Acceleration.} The inference speed of MLLMs has long been a concern in person ReID. However, recent advancements have introduced various methods to optimize the inference speed of MLLMs. We deployed Qwen3-VL-2B and Qwen3-VL-4B using vLLM and evaluated their Rank@1, mAP, and per-query inference time on RGB, text, and group person ReID tasks. As illustrated in \cref{tab:speed}, both models achieve a substantial increase in inference speed without any loss of performance. In particular, their inference speed on the text person ReID task is 9.18× and 7.81× faster, respectively. It is worth noting that all of these results were obtained using MLLMs in 16-bit precision on a single NVIDIA RTX 4090 GPU. This implies that applying quantization techniques or hardware could further accelerate inference speeds.

\noindent \textbf{Impact of Gallery Size.} To investigate the impact of gallery size on the RGB and text person ReID tasks, we performed experiments by varying the gallery size from 50 to 5,000. The results can be found in \cref{fig:bar}. We first observe that Qwen2.5-VL-7B and Qwen3-VL-4B experience a sharp performance drop as the gallery size increases to  500. Beyond this point, although the performance continues to decline with larger gallery sizes, the rate of degradation becomes significantly slower. In addition, both models maintain relatively stable performance on the text person ReID task, highlighting the strong vision–language cross-modal understanding capabilities of MLLMs. Finally, Qwen3-VL-4B demonstrates greater robustness to variations in gallery size compared with Qwen2.5-VL-7B. When the gallery size reaches 5,000, Qwen3-VL-4B achieves 0.40 in Rank@1 and 0.48 in mAP on the text person ReID task, which is approximately half of its performance when the gallery size is 50.

\begin{figure}[t]
  \centering
  \includegraphics[width=\linewidth]{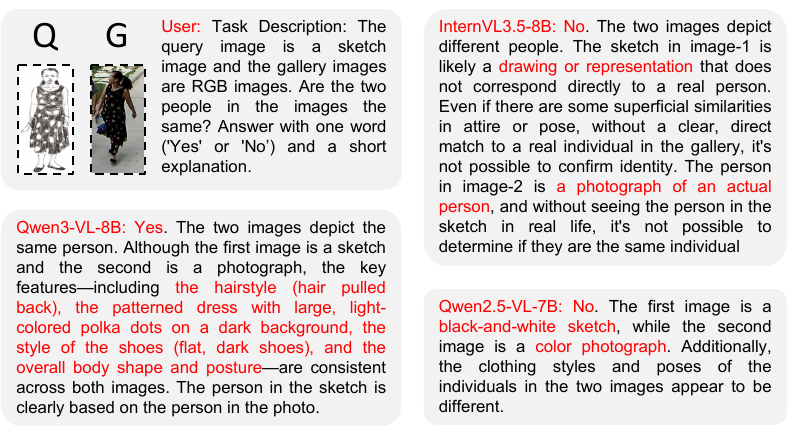}
  \caption{Qualitative comparisons of the sketch person ReID results from Qwen2.5-VL-7B, Qwen3-VL-8B, and InternVL3.5-8B.}
  \label{fig:examples}
\end{figure}

\noindent \textbf{Qualitative Comparisons.} Compared with traditional person ReID models, MLLMs offer stronger interpretability, allowing for clearer insights into their reasoning processes. We modify the original prompt to generate brief explanations for the answers provided by MLLMs. The experiments are conducted on the sketch person ReID task, which is widely adopted in forensic applications but remains highly challenging due to the unsatisfactory performance of most models. The results are presented in \cref{fig:examples}. Given a sketch image, Qwen3-VL-8B precisely captures fine-grained features such as hairstyle, dress patterns, shoe style, and overall body shape and posture. In contrast, Qwen2.5-VL-7B and InternVL3.5-8B primarily focus on modality differences and fail to extract high-level semantic information from these images, leading to incorrect results.

\subsection{Visualizations}

\begin{figure}[t]
  \centering
  \includegraphics[width=\linewidth]{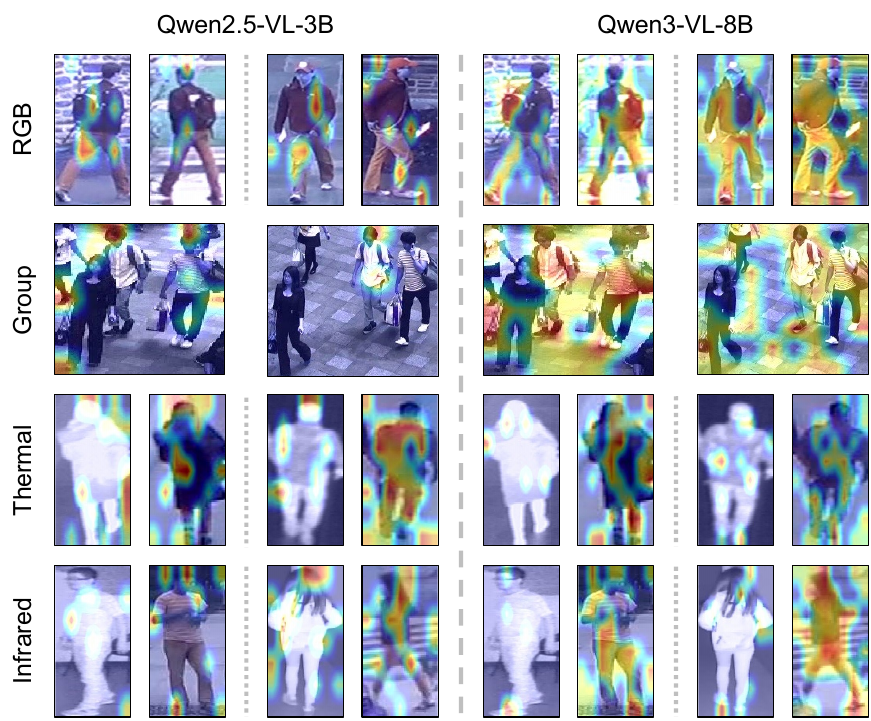}
  \caption{Attention maps of the RGB, group, thermal, and infrared person ReID results from Qwen2.5-VL-3B and Qwen3-VL-8B.}
  \label{fig:vis}
\end{figure}

To provide an intuitive understanding and further validate the effectiveness of MLLMs in person ReID, we conduct a qualitative analysis in which the visualization of attention maps from the MLLM language decoder is shown in \cref{fig:vis}. Specifically, we present examples from group, thermal, and infrared person ReID, each with two query images and two gallery images (one for group ReID). The visualization results reveal that MLLMs are capable of effectively capturing high-level semantic information while simultaneously attending to fine-grained appearance details. For instance, Qwen3-VL-8B precisely identifies the key features that distinguish different individuals in the RGB person ReID task. In the first example, the model primarily focuses on the carried bag. Furthermore, when the bag is occluded in the second example, its attention shifts toward the subject's posture instead. In contrast, we observe that Qwen2.5-VL-3B fails to capture discriminative features between query and gallery images. This also explains the model's poor performance in thermal and infrared person ReID tasks, where limited attention is concentrated on the relevant regions.
\section{Conclusion}

In this paper, we propose VP-ReID, a versatile person ReID benchmark specifically designed for MLLMs. The VP-ReID includes 257,310 multi-modal queries and gallery images, covering two evaluation schemes and ten different person ReID tasks. Extensive experiments demonstrate that MLLMs achieve impressive performance in various person ReID tasks. However, they also struggle with challenging thermal and infrared modalities, which require comprehensive cross-modal understanding capabilities and professional domain knowledge. Additionally, our findings reveal that some families of MLLMs adhere to the scaling law in person ReID tasks, exhibiting correlated performance when dealing with similar modalities. We hope these results can provide insights into the future improvements of MLLMs for effective and versatile person ReID.

\clearpage  
\appendix
\clearpage
\setcounter{page}{1}

\maketitlesupplementary

\section{Real-world Demonstration}

\begin{figure*}[b]
  \centering
  \includegraphics[width=\linewidth]{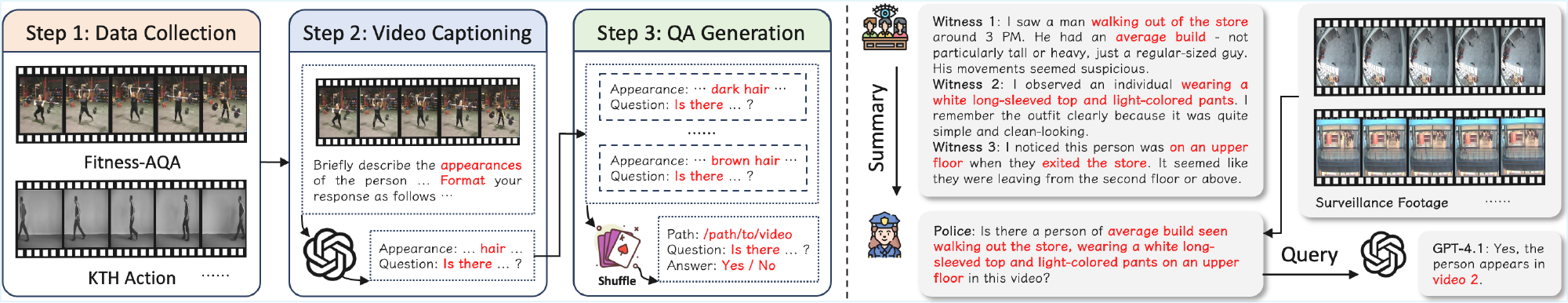}
  \caption{Overview of the real-world demonstration. Left: the construction pipeline of the test set used in the real-world demonstration. Right: a simulation case illustrating forensic applications.}
  \label{fig:overview}
\end{figure*}

Existing video-based person ReID datasets predominantly consist of pre-cropped pedestrian tracks, which exhibit considerable disparity from raw surveillance footage in real-world scenarios. To facilitate the real-world applications of MLLM-based person ReID methods, we propose a video-based person ReID dataset that maintains a similar data structure to MMReID-Bench. The detailed construction pipeline is illustrated in \cref{fig:overview}. We first collect source data from three real-world video datasets: Fitness-AQA \cite{parmar2022domain}, CAVIAR \cite{cheng2011custom}, and KTH Action \cite{roth2009efficient}. Then, the collected data is captioned by GPT-4.1 in a structured format. Finally, question-answer pairs are generated by shuffling and matching the videos and captions.

As shown in \cref{fig:overview}, the inputs of our demonstration consist of a video clip and a textual description, which are widely used in forensic applications. We consider a scenario in which multiple witnesses provide different descriptions of a suspect at a crime scene. The police summarize the descriptions and search for the suspect across the surveillance footage using MLLMs. Ultimately, the MLLMs analyze the videos and output the identified target.

Given their competitive performance and open-source availability, we adopt the Qwen2.5-VL family for our experiments. The results are presented in \cref{tab:demo}. It can be observed that even Qwen2.5-VL-7B achieves a notable F1 score of 0.812. We assume that this can be attributed to the fine-grained descriptions generated by GPT-4.1, which may not be available in real-world scenarios. However, as a case progresses, increasing amounts of evidence and witness testimonies will gradually accumulate. Consequently, the descriptions of the suspect will become more detailed and will closely resemble the conditions in our experimental setup. Therefore, the results of our experiments hold substantial practical relevance for real-world forensic applications.

\begin{table}[t]
    \centering
    \caption{Results of Qwen2.5-VL family on our real-world demonstration. The best and second-best results are in \textbf{bold} and \underline{underlined}, respectively.}
    \label{tab:demo}
    \begin{tabular}{@{}cccc@{}}
        \toprule
        Models & Precision$\uparrow$ & Recall$\uparrow$ & F1 score$\uparrow$ \\
        \midrule
        Qwen2.5-VL-3B & 0.640 & 0.300 & 0.408 \\
        Qwen2.5-VL-7B & \underline{0.763} & \underline{0.867} & \underline{0.812} \\
        Qwen2.5-VL-32B & 0.757 & \textbf{0.893} & \textbf{0.819} \\
        Qwen2.5-VL-72B & \textbf{0.787} & 0.800 & 0.793 \\
        \bottomrule
    \end{tabular}
\end{table}

\section{Detailed Prompts for the QGM}
\label{qgm}

In this section, we provide the prompts used for the QGM, as detailed in \cref{tab:qgm}. The placeholders for text and images need to be adjusted accordingly for different models.

\begin{table}[t]
\centering
\caption{Detailed Prompts for the QGM.}

\begin{tcolorbox}[
    colback=gray!10,
    colframe=black,
    boxrule=0.5pt,
    width=\linewidth,
    arc=2pt,
    left=6pt,
    right=6pt,
    top=6pt,
    bottom=6pt
]

\textbf{Prompt for the Image-based (Exclude Group) Person ReID:}

You are a person re-identification assistant. Are the two people in the images the same? Answer strictly with only one word: `Yes' or `No'.

\textbf{Prompt for the Group ReID:}

You are a person re-identification assistant. Are the two group of people in the images the same? Answer strictly with only one word: `Yes' or `No'.

\textbf{Prompt for the Text Person ReID:}

You are a person re-identification assistant. Your task is to carefully analyze the descriptions of the person in the following text and determine whether the image shows the same person.\textbackslash $<$text$>$ \textbackslash strictly with only one word: `Yes' or `No'.

\end{tcolorbox}
\label{tab:qgm}
\end{table}

\section{Detailed Prompts for the MCQ}
\label{sec:mcq}

In this section, we provide the prompts used for the MCQ, as detailed in \cref{tab:mcq_prompt}. The placeholders for text and images need to be adjusted accordingly for different models.

\begin{table*}[t]
\fontsize{9pt}{11pt}\selectfont
\centering
\caption{Detailed Prompts for the MCQ.}

\begin{tcolorbox}[
    colback=gray!10,
    colframe=black,
    boxrule=0.5pt,
    width=\linewidth,
    arc=2pt,
    left=6pt,
    right=6pt,
    top=6pt,
    bottom=6pt
]

\textbf{Prompt for the Cloth-changing Person ReID:}

Query Image: $<\texttt{image 1}>$ \textbackslash n Gallery Images: $<\texttt{image 1}>$ $<\texttt{image 2}>$ $<\texttt{image 3}>$ $<\texttt{image 4}>$ \textbackslash n Task Description: The people in the query image and gallery images wear different clothes. Your task is to carefully analyze the appearance, and pose of the person in the query image and determine which gallery image shows the same person. Please answer with the letter of the correct option directly (A, B, C, or D) and no other text.

\textbf{Prompt for the RGB Person ReID:}

Query Image: $<\texttt{image 1}>$ \textbackslash n Gallery Images: $<\texttt{image 1}>$ $<\texttt{image 2}>$ $<\texttt{image 3}>$ $<\texttt{image 4}>$ \textbackslash n Task Description: The query and gallery images are all RGB images. Your task is to carefully analyze the appearance, clothing, and pose of the person in the query image and determine which gallery image shows the same person. Please answer with the letter of the correct option directly (A, B, C, or D) and no other text.

\textbf{Prompt for the Occluded Person ReID:}

Query Image: $<\texttt{image 1}>$ \textbackslash n Gallery Images: $<\texttt{image 1}>$ $<\texttt{image 2}>$ $<\texttt{image 3}>$ $<\texttt{image 4}>$ \textbackslash n Task Description: The query image is a thermal image and the gallery images are RGB images. Your task is to carefully analyze the body structure, contour features, thermal pattern, gait/motion pattern, background information, and other details of the person in the query image and determine which gallery image shows the same person. Please answer with the letter of the correct option directly (A, B, C, or D) and no other text.

\textbf{Prompt for the Synthetic Person ReID:}

Query Image: $<\texttt{image 1}>$ \textbackslash n Gallery Images: $<\texttt{image 1}>$ $<\texttt{image 2}>$ $<\texttt{image 3}>$ $<\texttt{image 4}>$ \textbackslash n Task Description: The query and gallery images are all synthetic. Your task is to carefully analyze the appearance, clothing, and pose of the person in the query image and determine which gallery image shows the same person. Please answer with the letter of the correct option directly (A, B, C, or D) and no other text.

\textbf{Prompt for the Sketch Person ReID:}

Query Image: $<\texttt{image 1}>$ \textbackslash n Gallery Images: $<\texttt{image 1}>$ $<\texttt{image 2}>$ $<\texttt{image 3}>$ $<\texttt{image 4}>$ \textbackslash n Task Description: The query image is a sketch image and the gallery images are RGB images. Your task is to carefully analyze the appearance, clothing, and pose of the person in the query image and determine which gallery image shows the same person. Please answer with the letter of the correct option directly (A, B, C, or D) and no other text.

\textbf{Prompt for the Thermal ReID:}

Query Image: $<\texttt{image 1}>$ \textbackslash n Gallery Images: $<\texttt{image 1}>$ $<\texttt{image 2}>$ $<\texttt{image 3}>$ $<\texttt{image 4}>$ \textbackslash n Task Description: The query image is a thermal image and the gallery images are RGB images. Your task is to carefully analyze the body structure, contour features, thermal pattern, gait/motion pattern, background information, and other details of the person in the query image and determine which gallery image shows the same person. Please answer with the letter of the correct option directly (A, B, C, or D) and no other text.

\textbf{Prompt for the Text Person ReID:}

Query Image: $<\texttt{text 1}>$ \textbackslash n Gallery Images: $<\texttt{image 1}>$ $<\texttt{image 2}>$ $<\texttt{image 3}>$ $<\texttt{image 4}>$ \textbackslash n Task Description: The query text contains 2 textual descriptions about the attributes of the target person. Your task is to carefully analyze the descriptions of the person in the query text and determine which gallery image shows the same person. Please answer with the letter of the correct option directly (A, B, C, or D) and no other text.

\textbf{Prompt for the Group ReID:}

Query Image: $<\texttt{image 1}>$ \textbackslash n Gallery Images: $<\texttt{image 1}>$ $<\texttt{image 2}>$ $<\texttt{image 3}>$ $<\texttt{image 4}>$ \textbackslash n Task Description: The query and gallery images are all RGB images. Your task is to carefully analyze the relative position, motion pattern, structure topology, interaction features, individual appearance, size, and density of the people in the query image and determine which gallery image shows the same people. Please answer with the letter of the correct option directly (A, B, C, or D) and no other text.

\textbf{Prompt for the Infrared ReID:}

Query Image: $<\texttt{image 1}>$ \textbackslash n Gallery Images: $<\texttt{image 1}>$ $<\texttt{image 2}>$ $<\texttt{image 3}>$ $<\texttt{image 4}>$ \textbackslash n Task Description: The query image is an infrared image and the gallery images are RGB images. Your task is to carefully analyze the body structure, contour features, thermal pattern, gait/motion pattern, background information, and other details of the person in the query image and determine which gallery image shows the same person. Please answer with the letter of the correct option directly (A, B, C, or D) and no other text.

\textbf{Prompt for the UAV Person ReID:}

Query Image: $<\texttt{image 1}>$ \textbackslash n Gallery Images: $<\texttt{image 1}>$ $<\texttt{image 2}>$ $<\texttt{image 3}>$ $<\texttt{image 4}>$ \textbackslash n Task Description: The query and gallery images are all UAV images. Your task is to carefully analyze the appearance, body structure, contour features, gait/motion pattern, background information, and other details of the person in the query image and determine which gallery image shows the same person. Please answer with the letter of the correct option directly (A, B, C, or D) and no other text.

\end{tcolorbox}
\label{tab:mcq_prompt}
\end{table*}

\section{More Inference Examples}

\begin{figure*}[t]
  \centering
  \includegraphics[width=0.93\linewidth]{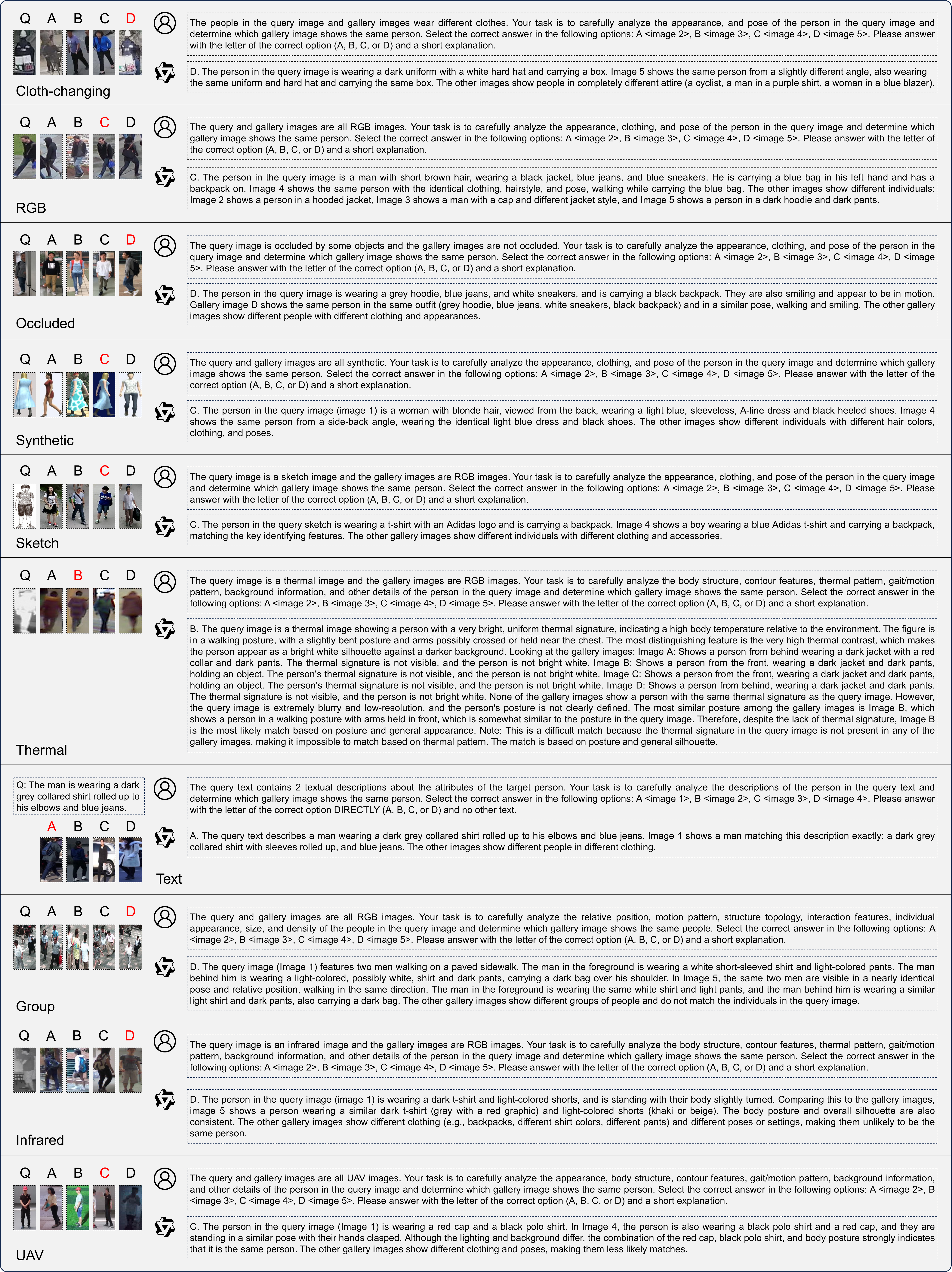}
  \caption{More inference examples from Qwen3-VL-8B in the MCQ. We highlight the correct answers in red.}
  \label{fig:inference}
\end{figure*}

We show the results of Qwen3-VL-8B on 10 person ReID tasks in \cref{fig:inference}. Note that we modify the original prompt to generate a short explanation.

\clearpage 

{
    \small
    \bibliographystyle{ieeenat_fullname}

}

\end{document}